\documentclass{article}
\usepackage[final]{corl_2026} % initial submission (anonymised). Use [final] or [preprint] later.
\makeatletter
\renewcommand{\@noticestring}{Accepted at HumanRLI Workshop @ CoRL 2026.}
\makeatother
\usepackage[utf8]{inputenc}
\usepackage[T1]{fontenc}
\usepackage{amsmath,amssymb}
\usepackage{graphicx}
\usepackage{booktabs}
\usepackage{multirow}
\usepackage{placeins}

\usepackage{titlesec}
\titlespacing*{\section}{0pt}{8pt plus 2pt minus 2pt}{4pt plus 1pt minus 1pt}
\titlespacing*{\subsection}{0pt}{6pt plus 2pt minus 2pt}{3pt plus 1pt minus 1pt}
\titlespacing*{\paragraph}{0pt}{4pt plus 1pt minus 1pt}{0.8em}
\usepackage{enumitem}
\setlist{itemsep=1pt,topsep=2pt,parsep=0pt,partopsep=0pt}
\AtBeginDocument{%
  \setlength{\abovedisplayskip}{4pt plus 1pt minus 1pt}%
  \setlength{\belowdisplayskip}{4pt plus 1pt minus 1pt}%
  \setlength{\abovedisplayshortskip}{2pt}%
  \setlength{\belowdisplayshortskip}{2pt}}

\newcommand{\piz}{\pi_0}
\newcommand{\pimu}{\pi_{-u}}
\newcommand{\best}[1]{\textbf{#1}}
\newcommand{\fig}[2]{\IfFileExists{#2}{\includegraphics[width=#1]{#2}}{\fbox{\parbox[c][3cm][c]{#1}{\centering\small missing: \texttt{\detokenize{#2}}}}}}

\title{Who Taught the Robot That? Removing One User's Corrections from a Robot Policy That Learns Online}

\author{
  Eshika Pathak\\
  University of Illinois Urbana-Champaign\\
  \texttt{epathak2@illinois.edu}
  \And
  Leela Krishna\\
  Centific\\
  \texttt{leela.krishna@centific.com}
}

\begin{document}
\maketitle

\begin{abstract}
A robot that keeps learning after deployment is corrected by several people in turn, and one of them may later ask to be removed. We show that whether the robot can forget that person depends on how people correct it. The goal of removal is the robot that would exist if the person had never been there, which we compute exactly in two simulated tasks with scripted users. When users correct the robot by taking over control, an ordered log of training updates is enough: retraining on the remaining sessions in their original order gives a robot as close to the goal as training noise allows. Deleting the person's data and retraining in one pass does not, even when the other users' corrections do not depend on the removed person, because it loses the order of training. When users only nudge the robot's action, the other users' corrections carry the removed person's influence, and no method we tested recovers the goal robot. Continuing to train without the person removes most of their habits, but not what their data in the training buffer keep alive. The log can identify one clearly harmful user, but not rank users by influence better than counting their corrections.
\end{abstract}


\section{Introduction}

Picture a robot in a shared home or a small clinic. Several people correct it over weeks: one steers it away from the stairs, another changes where it sets things down, a third slows it down. After each person's corrections, the robot's control policy, the function that chooses its action in each situation, is retrained. Later one of these people has to be removed. They withdraw consent, they move out, or their corrections were wrong and the robot now fails at its task. The operator needs to undo what that person taught the robot.

Machine unlearning studies how to remove one contributor's data from a trained model. In federated learning, methods such as FedEraser~\citep{liu2021federaser} keep a log of every contributor's past model updates and use it to rebuild the model without one of them~\citep{romandini2024federated,liu2024survey}. These methods assume that each contributor's data are fixed and do not depend on the other contributors. Robot corrections break this assumption. A person corrects the robot while it acts, in situations the robot reached because of everyone who corrected it before. If user A teaches the robot to use the lower of two gaps in a wall, user B later corrects a robot that already heads for the lower gap, so B's corrections reflect what A taught. Deleting A's data may therefore leave some of A's influence in the robot.

We study this question in simulation, where we can compute the robot that never met the removed user by rerunning the whole deployment without them. Our main finding is that whether a robot can forget one person's corrections depends on how people correct it. If people take over control, their influence can be removed using an ordered log of training updates. If people only nudge the robot's action, it cannot be removed with any method we tested. In detail:
\begin{enumerate}
  \item \textbf{Deleting a user's data and retraining on the rest in one pass gives the wrong robot, mainly because it loses the order of training.} It fails by the same amount when we make the users' corrections independent of each other by design. Retraining on the remaining sessions in their original order gives the right robot, up to random training noise; the same sessions in shuffled order do not.
  \item \textbf{When users only nudge the robot, no method we tested gives the right robot.} A nudge contains part of the robot's own action, which the removed user had shaped. For nudges that move the action 20 to 60\% of the way toward the user's choice, retraining in order stays farther from the right robot than training noise explains, and for nudges up to 45\% it keeps too much of the removed user's habits. Rebuilding the corrections from the log fixes part of this, but not all, because the removed user also changed which situations the others corrected.
  \item \textbf{Continuing to train without the user removes most of their habits, except what their old data keep alive.} A habit that all remaining users oppose shrinks to 15\% of its initial size and stays there, because the removed user's corrections are still reused in training. Deleting those corrections, without any other change, shrinks it to 2\%.
  \item \textbf{The log does not reliably say who taught the robot a behaviour.} It identifies one clearly harmful user among three in 85 to 95\% of runs. Among six users with random preferences, its ranking of who influenced a behaviour most agrees only weakly with the true ranking, and no better than simply counting each user's corrections.
\end{enumerate}
A 2D delivery task and a simulated robot arm give the same answers for the two correction types we ran on both (findings 1 and 2).

\section{Related work}

\paragraph{Machine unlearning.} FedEraser~\citep{liu2021federaser} stores each contributor's past updates in federated learning and uses them to remove a contributor; surveys cover the wider field~\citep{romandini2024federated,liu2024survey,zhao2024exploring}. SISA~\citep{bourtoule2021sisa} makes retraining exact by saving model checkpoints and retraining only from the last checkpoint before the removed data. Our replay method is exact retraining from a checkpoint in this sense. We add a goal for removal when data are collected interactively, a test of when replay reaches it, and the finding that training order matters even for independent data.

\paragraph{Unlearning in robot policies.} \citet{anon2026rethinking} evaluate removal of demonstrations from imitation-learning policies against a policy retrained without them, and measure both the policy's behaviour and statistical traces of the removed data. VLA-Forget~\citep{anon2026vlaforget} and TrajDeleter~\citep{gong2025trajdeleter} remove data from a fixed dataset, and MoRE~\citep{anon2026uncloning} suppresses unwanted behaviours by editing the network's weights. All of these work with a fixed dataset; we study a robot that learns from several people over time, and compare against a rerun of the deployment rather than a retrained policy.

\paragraph{Learning from interventions and shared control.} Our takeover users follow HG-DAgger~\citep{kelly2019hgdagger}, in which a person takes control when the robot errs and the robot learns from what the person did. Our partial users follow policy blending in shared control~\citep{dragan2013policy}, in which the executed action is a mix of the robot's and the person's. Data attribution~\citep{agia2025cupid,anon2026qoq} and multi-user preference learning~\citep{anon2026prec,civit2025argumentation} do not consider removing a person after deployment.

\section{Setting}

\subsection{How the robot learns from users}

The robot starts with a base policy $\piz$, trained by imitating 200 demonstrations. It is then deployed, and users take turns in \emph{sessions}. In a session, one user watches the robot for 10 \emph{episodes} (attempts at the task). When the robot does something the user dislikes, the user intervenes, and the actions the user supplies are recorded as \emph{corrections}: pairs of a state and the action the robot should have taken there. After the 10 episodes, the policy is fine-tuned for 200 gradient steps. Each training batch is drawn 50\% from this session's corrections, 25\% from the corrections of all earlier sessions (we call these stored earlier corrections the \emph{replay buffer}) and 25\% from the original demonstrations.

Three users, A, B and C, each get four sessions. A \emph{round} gives each user one session. The schedule's \emph{block length} $k$ sets how often users alternate: with $k{=}1$ the order is A\,B\,C\,A\,B\,C\,\ldots; with $k{=}4$ it is AAAA\,BBBB\,CCCC.

\subsection{What removing a user should produce}

To remove user $u$, the goal is the policy the robot would have if $u$ had never taken part. We call it the \emph{counterfactual policy} $\pimu$ and compute it by rerunning the whole deployment without $u$'s sessions, using the same random numbers for every other session. This is not the same as training on everyone's data except $u$'s: without $u$, the other users would have seen a different robot, intervened at different moments and given different corrections. A real operator cannot compute $\pimu$, because they cannot ask the other users to repeat weeks of corrections. Scripted users let us compute it, so we can measure how close each removal method gets.

\subsection{Two ways of correcting}

Each user has a preferred \emph{expert action} $a^{\mathrm{exp}}$ in every state. A \emph{takeover} user, at the first moment the robot does something they dislike, takes control and drives the robot with $a^{\mathrm{exp}}$ until the end of the episode. A \emph{partial} user, from that moment on, only nudges the robot: the executed and recorded action is
\begin{equation*}
  a = (1-\alpha)\, a^{\mathrm{robot}} + \alpha\, a^{\mathrm{exp}}, \qquad 0 < \alpha \le 1,
\end{equation*}
where $a^{\mathrm{robot}}$ is what the robot would have done and $\alpha$, the \emph{correction strength}, is how far the user moves the action toward their own choice. At $\alpha{=}1$ the partial user behaves exactly like a takeover user. At $\alpha{=}0.3$, the recorded action is 70\% the robot's own action. Since the robot's action reflects what earlier users taught it, so does the recorded correction.

\subsection{Correction log}

After each session we store one log record: who corrected and when, the corrections themselves, the robot's own action at each correction, the random seed and number of steps used for the update, and a copy of the policy before the update (a checkpoint). Repeating the logged updates in order, starting from $\piz$, reproduces the deployed policy exactly.

\subsection{Removal methods}

\begin{description}
  \item[Retrain.] Start from $\piz$ and train once on all logged corrections except $u$'s, mixed together in random batches, for the same total number of steps. This is what ``delete the user's data and retrain'' usually means.
  \item[Replay.] Start from the checkpoint saved just before $u$'s first session, then repeat every later session except $u$'s, in the original order, with the same seeds and the same batch rule. The result is the same as replaying from $\piz$; the checkpoint only saves computation.
  \item[Shuffled replay.] Replay with the remaining sessions in random order. This tests whether the order matters.
  \item[Relabelled replay.] For partial users only. Replay, but rebuild each correction using the policy being replayed instead of the deployed policy: the new recorded action is the replayed policy's action plus the user's logged nudge, $\pi_{\mathrm{replay}}(s) + (a - a^{\mathrm{robot}})$. An \emph{oracle} version instead recomputes the correction from the user's true expert action. An operator cannot use the oracle, because they do not know each user's expert action; it shows the best that relabelling could do.
  \item[Other methods.] \emph{Gradient-ascent unlearning} starts from the deployed policy and trains it to fit $u$'s corrections worse while still fitting everyone else's. \emph{Per-user adapters} keep $\piz$ fixed and give each user a small add-on network (a LoRA adapter) trained only on their sessions; removal deletes $u$'s adapter. \emph{Continuing without the user} keeps the deployed policy and goes on training with the remaining users.
\end{description}

\subsection{How we measure success}

\textbf{Action distance.} To compare two policies we give them the same states and measure how differently they act: $d(\pi_1, \pi_2)$ is the average Euclidean distance between their actions, over 500 states visited by $\pimu$ and 500 states where $u$ corrected the robot. The units are those of the action (in Delivery2D, a velocity per step, at most 0.05).

\textbf{Noise floor.} Training is random: two runs of the same deployment with different random seeds end with slightly different policies. We measure this by rerunning $\pimu$ with three new training seeds and averaging their distance to $\pimu$. We call this the \emph{noise floor}. A removal method whose distance to $\pimu$ equals the noise floor is as close as one can expect; we often report a distance as a multiple of the floor, so that 1 means ``as close as training noise allows''. Replay repeats the sessions before $u$'s first session exactly as $\pimu$ did, which removes some of the randomness. For replay we therefore use a \emph{prefix floor}, computed by rerunning $\pimu$ with new seeds only from $u$'s first session onward. The floor differs between experimental conditions, so when we compare conditions we use raw distances.

\textbf{Share of influence left.} To show how much of $u$'s influence a method removes, we divide its distance to $\pimu$ by the distance of the deployed policy, which still contains all of $u$'s influence:
\begin{equation*}
  r = \frac{d(\pi_{\mathrm{method}}, \pimu)}{d(\pi_{\mathrm{deployed}}, \pimu)}, \qquad
  r_{\min} = \frac{\text{noise floor}}{d(\pi_{\mathrm{deployed}}, \pimu)} .
\end{equation*}
$r{=}1$ means the method removed nothing. $r$ cannot fall much below $r_{\min}$, because even a perfect method stays one noise floor away from $\pimu$. We report $r - r_{\min}$: 0 means the method did as well as possible.

\textbf{Preference match.} The share of 100 test episodes in which the robot does what user $u$ prefers: it passes the obstacle the way $u$ likes and places the object within 0.08 of $u$'s preferred position. We report a method's preference match minus that of $\pimu$. A positive value means the method keeps more of $u$'s habits than the counterfactual robot would have; a negative value means it removes more.

\textbf{Statistics.} Every comparison is made seed by seed: both methods are run on the same 20 seeds, and we report the mean difference, a 95\% bootstrap confidence interval and the $p$-value of a Wilcoxon signed-rank test. When we claim that two quantities are equal, we use a paired equivalence test (two one-sided tests) with the margin stated.

\section{Tasks}

\begin{figure}[t]
  \centering
  \fig{0.58\linewidth}{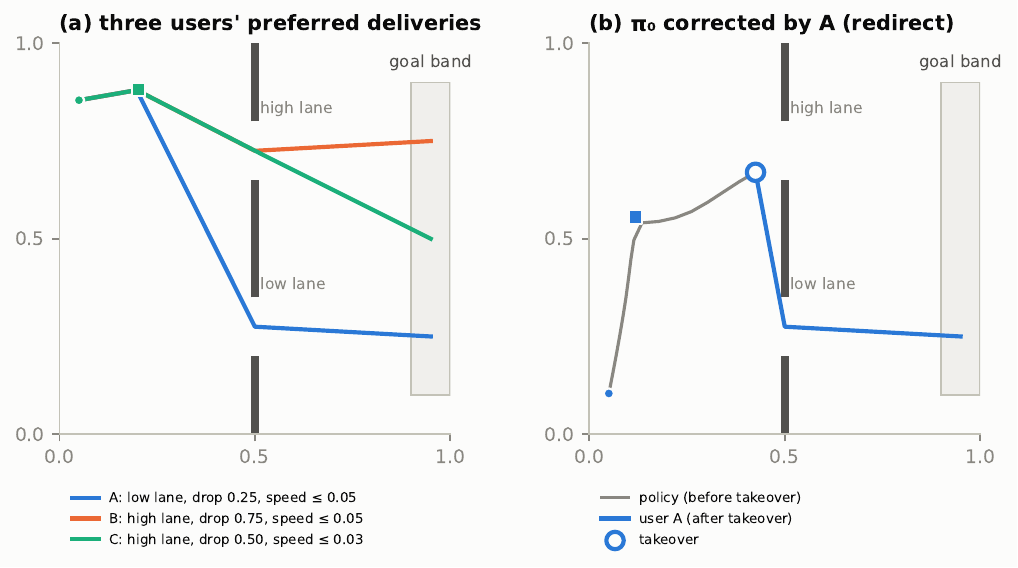}\\[2pt]
  \fig{0.58\linewidth}{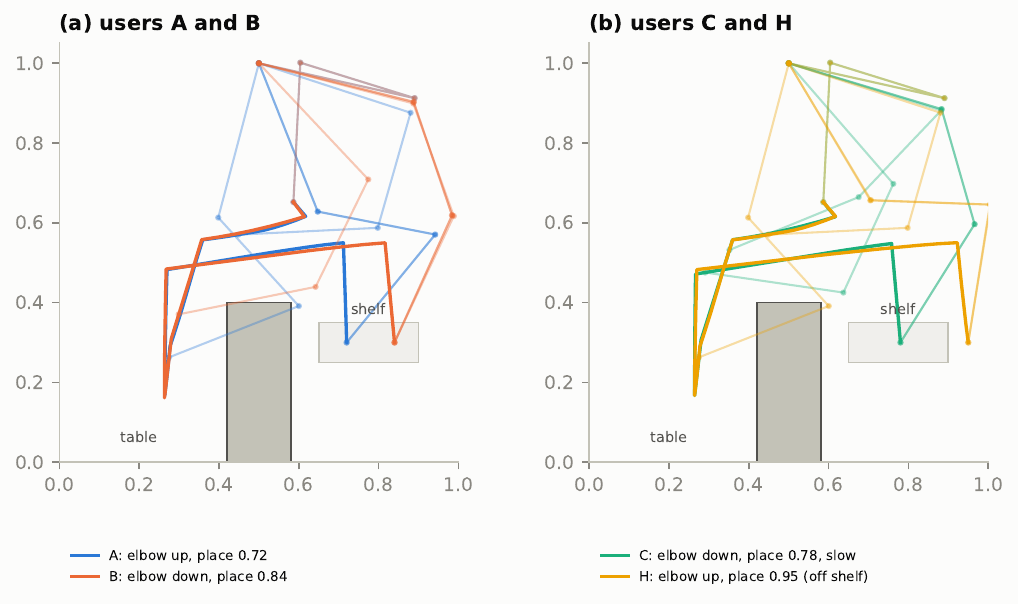}
  \caption{\textbf{The two tasks.} Top: Delivery2D, seen from above. (a) The robot picks up an object on the left, passes through the low or the high gap in a wall, and puts the object down in the goal band on the right. Users differ in which gap they prefer, where in the band they want the object, and how fast the robot may move. (b) The base policy heads for the wrong gap; user A takes over (circle) and finishes the episode their way. Bottom: Arm3, a planar arm with three joints. Thick lines show each user's preferred path of the hand; thin lines show the arm at several moments. The arm picks up an object from the table, carries it over the central block and places it on the shelf. Users differ in whether the elbow points up or down while carrying, where on the shelf the object goes, and how fast the joints may move. User H places the object off the shelf.}
  \label{fig:tasks}
\end{figure}

\textbf{Delivery2D} (Fig.~\ref{fig:tasks}, top). The state is the robot's position, the object's position and the goal's position. The action is a velocity, at most 0.05 per step, and an episode lasts 150 steps. A wall in the middle has a low gap and a high gap. User A prefers the low gap and a low drop position; B the high gap and a high drop; C the high gap, a middle drop and a slow speed. User H prefers a drop position just outside the goal band, so a robot that follows H fails the task. A user intervenes when the robot heads for the gap they do not want, puts the object down too far from their preferred position, moves faster than they allow, or drives away from the object before picking it up.

\textbf{Arm3} (Fig.~\ref{fig:tasks}, bottom). The arm moves in a vertical plane, and the action is a velocity for each of its three joints. The preferences are the elbow direction while carrying (the counterpart of the gap in Delivery2D), the position on the shelf, and a joint speed limit. The scripted users compute their actions with standard inverse-kinematics control and succeed in 98 to 100\% of episodes.

In both tasks the policy is a neural network with two hidden layers of 256 units. The base policy $\piz$ succeeds in 90\% of episodes on Delivery2D (between 52\% and 100\% depending on the seed) and 92\% on Arm3. We report all seeds; restricting the results to seeds where $\piz$ succeeds at least 80\% of the time changes none of them noticeably (Appendix~\ref{app:seeds}). Unless stated, results use 20 seeds; the Arm3 decomposition in Section~\ref{sec:order} and the experiments in Section~\ref{sec:fading} use 10.

\section{Results}

\subsection{With takeover users, one-pass retraining fails because it loses the order of training}
\label{sec:order}

Retrain and replay use the same logged corrections and differ only in how they train on them: retrain mixes everything in one pass, replay repeats the sessions in order. This lets us split retrain's distance from the counterfactual policy into two parts:
\begin{equation*}
  \underbrace{d(\text{retrain}, \pimu)}_{\text{total gap}}, \qquad
  \underbrace{d(\text{retrain}, \text{replay})}_{\text{due to the training procedure}}, \qquad
  \underbrace{d(\text{replay}, \pimu)}_{\text{due to the data}} .
\end{equation*}
The last part is what remains when the procedure is right; it measures how much the other users' corrections still carry the removed user's influence.

To test directly whether that influence is the problem, we also run a \emph{frozen-robot control}. In every session, the user corrects the base policy $\piz$ instead of the current, already-corrected robot; training is unchanged. No user then sees what earlier users taught, so no user's corrections depend on anyone else's. In this control the logged data are exactly the data the counterfactual run would collect, and replay equals $\pimu$ exactly. If retrain still fails here, the dependence between users is not the cause.

\begin{table}[t]
  \centering\small
  \caption{\textbf{Main results.} Each row is one task and one way of correcting, averaged over block lengths and over which user is removed. \emph{Retrain gap}: distance from retrain to the counterfactual policy $\pimu$, in action units ($\times 10^{-3}$), in the normal deployment and in the frozen-robot control, where no user's corrections depend on another's. \emph{Replay distance}: distance from replay to $\pimu$ as a multiple of the prefix noise floor (1 is as close as training noise allows). \emph{Preference}: the removed user's preference match minus that of $\pimu$ (0 is right; positive keeps too much of the user). $r - r_{\min}$: the share of the user's influence left beyond what training noise forces (0 is the best possible). Bold marks results within training noise of $\pimu$.}
  \label{tab:main}
  \setlength{\tabcolsep}{3pt}\footnotesize
  \begin{tabular}{llcccccc}
    \toprule
    & & \multicolumn{2}{c}{retrain gap ($\times 10^{-3}$)} & replay & \multicolumn{2}{c}{preference} & $r - r_{\min}$ \\
    \cmidrule(lr){3-4}\cmidrule(lr){6-7}
    task & correction & normal & control & distance & replay & retrain & replay / retrain \\
    \midrule
    \multirow{2}{*}{Delivery2D} & takeover        & 3.8 & 3.8 & \best{0.91} & \best{+0.00} & $-0.10$ & $-0.09$ / $+0.15$ \\
                                & partial, $\alpha{=}0.3$ & 4.7 & 2.3 & 1.50 & $+0.06$ & \best{+0.00} & $+0.23$ / $+0.25$ \\
    \midrule
    \multirow{2}{*}{Arm3}       & takeover        & 7.0 & 5.2 & \best{0.90} & \best{+0.00} & $-0.07$ & $-0.04$ / $+0.03$ \\
                                & partial, $\alpha{=}0.3$ & 8.5 & 3.5 & 1.69 & $+0.03$ & +0.01 & $+0.26$ / $+0.25$ \\
    \bottomrule
  \end{tabular}
\end{table}

\begin{figure}[t]
  \centering
  \fig{0.78\linewidth}{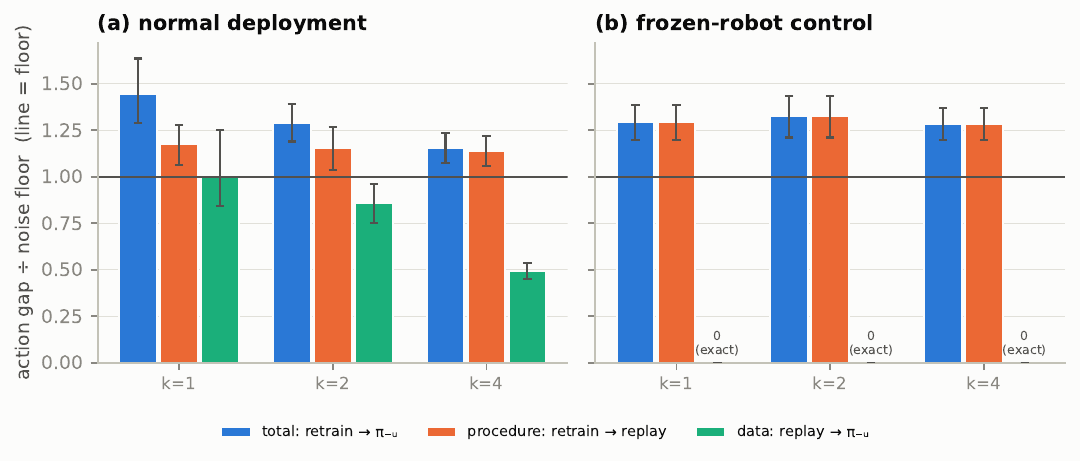}
  \caption{\textbf{Where retrain's error comes from,} Delivery2D, takeover users. Each group of bars is one schedule ($k$ is the block length). Blue: total distance from retrain to the counterfactual policy $\pimu$. Orange: distance from retrain to replay (due to the training procedure). Green: distance from replay to $\pimu$ (due to the data). All are divided by the noise floor; the horizontal line at 1 is as close as training noise allows. (a) Normal deployment. (b) Frozen-robot control, where no user's corrections depend on another's, so replay equals $\pimu$ exactly. Retrain's total error is the same in both panels.}
  \label{fig:decomp}
\end{figure}

\textbf{The dependence between users is not what makes retrain fail.} On Delivery2D, retrain ends 1.30 times the noise floor away from $\pimu$ in the normal deployment and also 1.30 times in the frozen-robot control (paired difference $-0.00$, 95\% CI $[-0.09, +0.09]$, $p{=}0.87$; Fig.~\ref{fig:decomp}, Table~\ref{tab:main}). On Arm3 the control lowers retrain's error by 26\%, so most of the error remains.

\textbf{Replay reaches the counterfactual policy.} Replay ends 0.91 times the prefix floor from $\pimu$ on Delivery2D and 0.90 on Arm3: as close as training noise allows (equal to the floor within $\pm 0.25$, $p{=}0.001$ on Delivery2D). It leaves the removed user's preference match where $\pimu$ has it ($+0.001$, equal within $\pm 0.05$), and it removes as much of the user's influence as noise allows ($r - r_{\min} \le 0$).

\textbf{What retrain loses is the order.} Replaying the same sessions in shuffled order moves the result 0.44 noise floors farther from $\pimu$ ($p{<}10^{-5}$), and 1.28 floors farther in the frozen-robot control. Two attempts to fix retrain without replaying in order, matching the deployment's mix of data in each batch and weighting each session by how often it was used, did not help (Appendix~\ref{app:retrainfix}). What retrain loses is not how often each session is seen, but that the late sessions are trained on last.

\textbf{Retrain's error is not specific to the removed user.} Whichever user is removed, retrain moves the robot toward the preferences of the user who comes first in each round and away from the others. When we reverse the order within each round from A\,B\,C to C\,B\,A, the favoured user changes from A ($+0.25$ preference match) to C ($+0.19$). This order effect also explains why retrain seems to remove too much of the removed user in Table~\ref{tab:main}: with B removed, B's preference match under retrain is $0.20$ below $\pimu$'s in one order and $0.23$ above it in the other.

\textbf{The dependence exists but is small.} When $u$ is present, the other users intervene for somewhat different reasons and in different places, but not more often, and the difference disappears when users do not alternate ($k{=}4$). Its one visible consequence is that retrain's error grows with alternation in the normal deployment but not in the control (Appendix~\ref{app:dependence}).

\subsection{With partial corrections, no method we tested reaches the counterfactual policy}
\label{sec:partial}

\begin{figure}[t]
  \centering
  \fig{\linewidth}{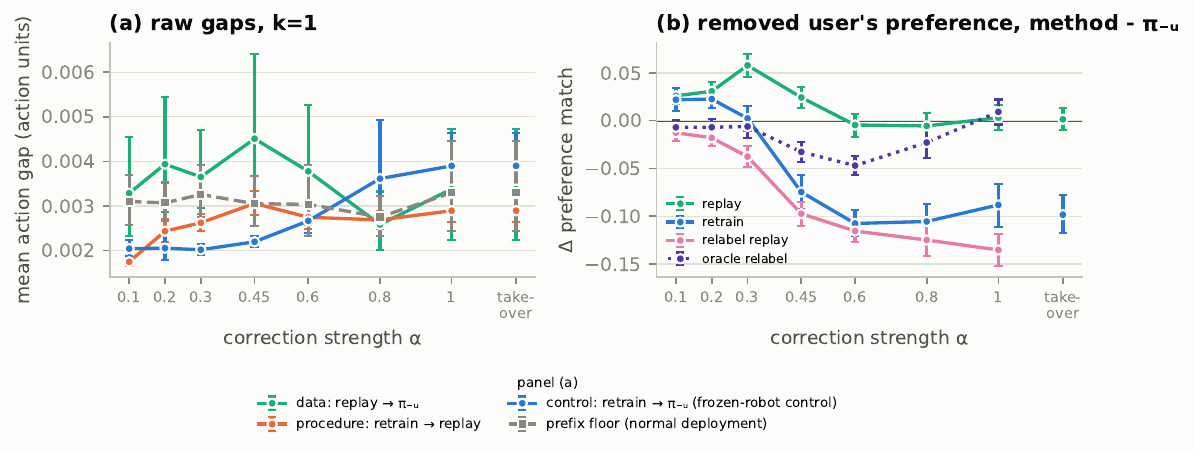}
  \caption{\textbf{Effect of correction strength $\alpha$,} Delivery2D, schedule $k{=}1$. ``Take-over'' is the takeover user, identical to $\alpha{=}1$. (a) Distances in action units. Green: replay to $\pimu$ (due to the data). Orange: retrain to replay (due to the training procedure). Blue: retrain to $\pimu$ in the frozen-robot control. Grey: prefix noise floor. (b) The removed user's preference match minus that of $\pimu$ for each method; 0 is right, and positive values keep too much of the user.}
  \label{fig:alpha}
\end{figure}

We now vary the correction strength $\alpha$ over 0.1, 0.2, 0.3, 0.45, 0.6, 0.8 and 1 on Delivery2D (Fig.~\ref{fig:alpha}).

\textbf{For intermediate correction strengths, the dependence between users matters.} For $\alpha$ from 0.1 to 0.6, retrain's error in the normal deployment is larger than in the frozen-robot control by $1.4$ to $3.1 \times 10^{-3}$ action units ($p{<}10^{-5}$ at each $\alpha$). For $\alpha \ge 0.8$ the difference is not significant. The effect is not symmetric: it is present at $\alpha{=}0.2$ and absent at $\alpha{=}0.8$. What matters most is how much of the robot's own action stays in each correction ($1-\alpha$), more than how strongly the removed user changed the robot.

\textbf{Replay keeps too much of the removed user.} For $\alpha$ from 0.2 to 0.6, replay ends farther from $\pimu$ than the prefix floor ($p \le 0.03$), and for $\alpha \le 0.45$ it keeps more of the removed user's preferences than $\pimu$ does (preference match $+0.024$ to $+0.058$). The reason is that the other users' corrections contain the robot's own action, which carried the removed user's habit, and replay trains on those corrections as they were recorded. Retrain is no better: for every $\alpha \le 0.6$ its $r$ is about 1 or more, so it removes no more of the user's influence than doing nothing. Replay removes as much as noise allows at $\alpha{=}0.1$ and at $\alpha \ge 0.8$; in between it is 0.16 to 0.36 above that limit. Arm3 at $\alpha{=}0.3$ gives the same picture (Table~\ref{tab:main}).

\textbf{Rebuilding the corrections helps only in part.} The oracle relabelling, which knows each user's true expert action, removes 43 to 64\% of replay's preference error for $\alpha \le 0.3$, but changes replay's action distance to $\pimu$ by only $-13$ to $+8\%$. The relabelling an operator can compute is worse than plain replay at every $\alpha$: its rebuilt correction differs from the true one by $\alpha(\pi_{\mathrm{replay}}(s) - \pi_{\mathrm{deployed}}(s))$, which grows with $\alpha$. Since even perfect labels leave the action distance unchanged, the rest of the error comes from where the corrections were made. The removed user changed which situations the other users saw and corrected, and the log cannot supply the corrections the others would have given in the situations the counterfactual robot reaches.

\subsection{Other removal methods}

Gradient-ascent unlearning and deleting a user's adapter both damage the robot: on Arm3 they lower task success from 0.76 to 0.54 and to 0.02. A membership-inference test finds traces of users only on Arm3 with partial corrections, and replay and retrain remove them (Appendix~\ref{app:other}).

\subsection{Continuing without the user, and the replay buffer}
\label{sec:fading}

\begin{table}[t]
  \centering\small
  \caption{\textbf{How H's preferred gap fades} when training continues with the remaining users, Delivery2D, takeover users, all remaining users preferring the other gap, 10 seeds. Each entry is the robot's share of trips through H's preferred gap minus the same share for the counterfactual policy $\pimu$ (0 means none of H's habit is left), after $R$ further rounds without H.}
  \label{tab:fading}
  \footnotesize
  \begin{tabular}{lcccc}
    \toprule
    & $R{=}0$ & $R{=}2$ & $R{=}4$ & $R{=}8$ \\
    \midrule
    continue, H's corrections kept in the replay buffer & $+0.66$ & $+0.18$ & $+0.12$ & $+0.10$ \\
    continue, H's corrections deleted from the replay buffer & $+0.66$ & $+0.02$ & $-0.00$ & $+0.01$ \\
    \bottomrule
  \end{tabular}
\end{table}

In a deployment with users A, B and the harmful user H, the robot ends up failing its task often: on Arm3 it succeeds in 33\% of episodes, against 60\% for the robot that never met H. An operator might simply stop listening to H and keep training with the others. H's harmful habit, putting the object down outside the goal, then fades quickly, because every remaining user corrects it each time it happens: on Delivery2D it returns to the level of $\pimu$ within 8 rounds, and on Arm3 two rounds bring it to 7\% of episodes, against 15\% for $\pimu$.

H's other habit, the choice of gap, does not fade fully on Delivery2D. When all remaining users prefer the other gap, H's habit falls from $+0.66$ to $+0.18$ in two rounds and then stays near $+0.10$, 15\% of its initial size (Table~\ref{tab:fading}). It persists because H's old corrections remain in the replay buffer and are reused in every later batch. Deleting them from the buffer, and changing nothing else, lowers the habit to $+0.01$ after 8 rounds (paired difference $-0.09$, $p{=}0.02$). When the remaining users disagree among themselves about the gap, the habit fades fully with or without the deletion. Appendix~\ref{app:fading} reports how the result changes when users notice only some violations, and with partial users.

\subsection{Can the log say which user taught a behaviour?}
\label{sec:attribution}

An operator may also want to answer the question in our title: which user taught the robot a behaviour it now shows? We test this on Delivery2D with six users whose preferences are drawn at random, three sessions each, interleaved, for three groups of users and 20 seeds each. We look at three behaviours $b$: the average drop position, the share of trips through the low gap, and the average speed. The true influence of user $v$ on a behaviour is how much the behaviour changes when $v$ is removed by rerunning the deployment:
\begin{equation*}
  I_v = b(\pi_{\mathrm{deployed}}) - b(\pi_{-v}).
\end{equation*}
We compare four ways an operator could estimate $I_v$ from the log, described in Appendix~\ref{app:attribution}. The main one removes $v$ by replay and measures the change in the behaviour; the simplest counts each user's logged corrections that push the behaviour, which needs no training. We score each by the Spearman rank correlation between its ranking of the six users and their true ranking: 1 means the same order, 0 means no relation.

Removing each user by replay gives a mean rank correlation of only 0.25 with takeover users and 0.32 with partial users at $\alpha{=}0.3$, no better than counting corrections (0.24 and 0.25; $p{=}0.92$ and $0.15$), although counting needs no training and replay about 18{,}000 gradient steps. Replay is better only at naming the single most influential user (45 to 50\% of cases, against 17\% for counting and for chance). With one clearly harmful user among A, B and H, replay names H as the most influential user in 85 to 95\% of seeds. Details are in Appendix~\ref{app:attribution}.

\section{Limitations}
\label{sec:limits}

\textbf{Simulation and scripted users.} Our users are programs with fixed preferences, one correction strength and perfect consistency. Real people correct inconsistently, change their minds and mix both ways of correcting. The counterfactual policy exists only in simulation, so an operator with real users cannot check replay against it.

\textbf{Small effects and small policies.} The deployed policy is only 1.0 to 1.75 noise floors from $\pimu$, so our conclusions rest on paired tests over 20 seeds and on preference measures, not on large differences; the same noise limits the attribution results. Both policies are small networks with low-dimensional inputs, and on Arm3 learning from corrections lowers task success a lot (from 0.92 to 0.60 in the harmful-user setting). Whether larger policies keep traces of a user that behaviour does not show is untested (Appendix~\ref{app:other}).

\section{Conclusion}

Whether a robot can forget one person's corrections depends on how people correct it. Corrections that replace the robot's action can be undone with an ordered log of training updates; one-pass retraining cannot undo them, because it loses the order of training. Partial corrections, as in shared control, could not be undone by any method we tested, because the other people's corrections carry the removed person's influence. Continuing to train without the person is not removal either, because their old data in the replay buffer keep part of their habits alive.

In practice, the log should store the order and random seeds of the updates, not only who gave which corrections. Checkpoints are optional, since replaying from $\piz$ gives the same result. The log is itself a privacy risk: it and every checkpoint saved after a user's first session contain that user, so a removal request means deleting both once the replay is done.

\bibliography{refs}

@article{romandini2024federated,
  title   = {Federated Unlearning: A Survey on Methods, Design Guidelines, and Evaluation Metrics},
  author  = {Romandini, Nicol{\`o} and Mora, Alessio and Mazzocca, Carlo and Montanari, Rebecca and Bellavista, Paolo},
  journal = {IEEE Transactions on Neural Networks and Learning Systems},
  year    = {2024},
  note    = {arXiv:2401.05146}
}

@article{liu2024survey,
  title   = {A Survey on Federated Unlearning: Challenges, Methods, and Future Directions},
  author  = {Liu, Ziyao and Jiang, Yu and Shen, Jiyuan and Peng, Minyi and Lam, Kwok-Yan and Yuan, Xingliang and Liu, Xiaoning},
  journal = {ACM Computing Surveys},
  year    = {2024},
  note    = {arXiv:2310.20448}
}

@article{zhao2024exploring,
  title   = {Exploring Federated Unlearning: Review, Comparison, and Insights},
  author  = {Zhao, Yang and Yang, Jiaxi and Tao, Yiling and Wang, Lixu and Li, Xiaoxiao and Niyato, Dusit},
  journal = {arXiv preprint arXiv:2310.19218},
  year    = {2024}
}

@inproceedings{liu2021federaser,
  title     = {{FedEraser}: Enabling Efficient Client-Level Data Removal from Federated Learning Models},
  author    = {Liu, Gaoyang and Ma, Xiaoqiang and Yang, Yang and Wang, Chen and Liu, Jiangchuan},
  booktitle = {IEEE/ACM International Symposium on Quality of Service (IWQoS)},
  year      = {2021}
}

@article{anon2026rethinking,
  title   = {Rethinking Demonstration Unlearning in Imitation Learning for Robotics},
  author  = {Li, Jiazhuo and Zhang, Yu and Fei, Yiming and Dong, Kangkang and Zhu, Xiaojun and Liu, Houde and Tao, Jinze},
  journal = {arXiv preprint arXiv:2608.20784},
  year    = {2026}
}

@article{anon2026vlaforget,
  title   = {{VLA-Forget}: Vision-Language-Action Unlearning for Embodied Foundation Models},
  author  = {Ranjan, Ravi and Polyzou, Agoritsa},
  journal = {arXiv preprint arXiv:2604.03956},
  year    = {2026},
  note    = {Also in the KnowFM 2026 workshop proceedings}
}

@article{anon2026uncloning,
  title   = {Behavior Uncloning: Distilling Mode Redirection into Policy Weights without Inference-Time Steering},
  author  = {Wang, Hao and Lei, Jiuzhou and Li, Dayou and Liu, Bangya and Zheng, Minghui and Li, Manling and Zhang, Ruohan and Fan, Zhiwen},
  journal = {arXiv preprint arXiv:2606.29201},
  year    = {2026}
}

@inproceedings{agia2025cupid,
  title     = {{CUPID}: Curating Data your Robot Loves with Influence Functions},
  author    = {Agia, Christopher and Sinha, Rohan and Yang, Jingyun and Antonova, Rika and Pavone, Marco and Nishimura, Haruki and Itkina, Masha and Bohg, Jeannette},
  booktitle = {Conference on Robot Learning (CoRL), PMLR 305},
  year      = {2025}
}

@article{anon2026qoq,
  title   = {Quality over Quantity: Demonstration Curation via Influence Functions for Data-Centric Robot Learning},
  author  = {Lee, Haeone and Min, Taywon and Kim, Junsu and Kang, Sinjae and Liu, Fangchen and Pinto, Lerrel and Lee, Kimin},
  journal = {arXiv preprint arXiv:2603.09056},
  year    = {2026}
}

@article{anon2026prec,
  title   = {Deployable Human Preference Alignment in Robotics: Learning Representative Rewards from Diverse Human Preferences},
  author  = {Kim, Taehyung and Lee, Gwangmo and Chang, Minjun and Lim, Sunghyun and Choi, Jongeun},
  journal = {arXiv preprint arXiv:2607.12466},
  year    = {2026}
}

@article{civit2025argumentation,
  title   = {Multi-User Personalisation in Human-Robot Interaction: Resolving Preference Conflicts Using Gradual Argumentation},
  author  = {Civit, Aniol and Andriella, Antonio and Sierra, Carles and Aleny{\`a}, Guillem},
  journal = {arXiv preprint arXiv:2511.03576},
  year    = {2025}
}

@inproceedings{kelly2019hgdagger,
  title     = {{HG-DAgger}: Interactive Imitation Learning with Human Experts},
  author    = {Kelly, Michael and Sidrane, Chelsea and Driggs-Campbell, Katherine and Kochenderfer, Mykel J.},
  booktitle = {IEEE International Conference on Robotics and Automation (ICRA)},
  year      = {2019}
}

@inproceedings{yeom2018privacy,
  title     = {Privacy Risk in Machine Learning: Analyzing the Connection to Overfitting},
  author    = {Yeom, Samuel and Giacomelli, Irene and Fredrikson, Matt and Jha, Somesh},
  booktitle = {IEEE Computer Security Foundations Symposium (CSF)},
  year      = {2018}
}

@inproceedings{bourtoule2021sisa,
  title     = {Machine Unlearning},
  author    = {Bourtoule, Lucas and Chandrasekaran, Varun and Choquette-Choo, Christopher A. and Jia, Hengrui and Travers, Adelin and Zhang, Baiwu and Lie, David and Papernot, Nicolas},
  booktitle = {IEEE Symposium on Security and Privacy},
  year      = {2021}
}

@inproceedings{gong2025trajdeleter,
  title     = {{TrajDeleter}: Enabling Trajectory Forgetting in Offline Reinforcement Learning Agents},
  author    = {Gong, Chen and Li, Kecen and Yao, Jin and Wang, Tianhao},
  booktitle = {Network and Distributed System Security Symposium (NDSS)},
  year      = {2025}
}

@article{dragan2013policy,
  title   = {A policy-blending formalism for shared control},
  author  = {Dragan, Anca D. and Srinivasa, Siddhartha S.},
  journal = {The International Journal of Robotics Research},
  volume  = {32},
  number  = {7},
  pages   = {790--805},
  year    = {2013}
}

\clearpage
\appendix

\section{Other removal methods}
\label{app:other}

\textbf{Gradient-ascent unlearning} starts from the deployed policy and, for 200 steps, trains it to fit $u$'s corrections worse while still fitting everyone else's, with a bound that keeps the loss finite. On Arm3 with takeover users it ends 10 noise floors from $\pimu$ and lowers task success from 0.76 to 0.54. Nothing in its objective says how much of the user to remove, so it removes too much and damages the rest of the policy.

\textbf{Per-user adapters} keep $\piz$ fixed and give each user a small add-on network (a rank-4 LoRA adapter on both hidden layers) trained only during their sessions. Deleting one user's adapter lowers task success on Arm3 to 0.02, because each adapter was trained while the others were active and learned to work together with them. An earlier Delivery2D run with the same protocol gave the same ordering: 23 noise floors for gradient ascent, and success 0.14 after deleting an adapter.

\textbf{Membership inference.} A loss-threshold membership-inference test~\citep{yeom2018privacy} asks whether a policy fits a user's logged corrections more closely than fresh corrections from the same user, a sign that it was trained on them. Its score, the area under the ROC curve (AUC), is 0.5 when there is no trace of the user. On Delivery2D it is 0.49 to 0.51 for every policy, including the deployed one. On Arm3 with partial users at $\alpha{=}0.3$ it finds a trace: 0.59 for the removed user in the deployed policy, and 0.55 for H both in the deployed policy and after two more rounds without H. Retrain and replay bring it back to 0.50 to 0.52. With takeover users on Arm3 it is 0.49 to 0.51.

\section{How the other users' corrections change when the removed user is present}
\label{app:dependence}

We compare, session by session, the other users' corrections in the normal deployment (with $u$) and in the counterfactual run (without $u$), which share the same random numbers. As a noise reference we compare the counterfactual run with a rerun that uses new training seeds. Delivery2D, takeover users, 20 seeds.

\begin{itemize}
  \item \textbf{Why they intervene.} Each intervention has a reason: wrong gap, wrong drop position, too fast, or leaving the object behind. The share of each reason differs between the two runs by 0.055 more than noise, measured as total variation distance, which is half the sum of the absolute differences in the shares ($p{<}10^{-4}$).
  \item \textbf{Where they intervene.} The sets of states where interventions start differ by 0.010 more than noise in squared maximum mean discrepancy (MMD), a standard measure of how different two sets of points are, which is 0 when they come from the same distribution ($p{<}10^{-4}$).
  \item \textbf{How much they correct.} The number of corrections per session does not change beyond noise ($p{=}0.20$).
  \item \textbf{Effect of the schedule.} When users do not alternate ($k{=}4$), the difference in where they intervene disappears ($p{=}0.96$).
  \item \textbf{Effect on retrain.} Retrain's error grows with alternation, by 0.29 noise floors from $k{=}4$ to $k{=}1$ ($p{<}10^{-3}$), in the normal deployment but not in the frozen-robot control.
\end{itemize}

\section{Two attempts to fix retrain without replaying in order}
\label{app:retrainfix}

\textbf{Matching the batch mix.} During deployment, each batch is 50\% the current session, 25\% earlier sessions and 25\% demonstrations. A version of retrain whose batches have the same average mix, still in one pass, changed retrain's distance to $\pimu$ by $+0.08$ noise floors, so it did not help.

\textbf{Weighting each session by how often it was used.} We gave each session's data the weight it effectively had during deployment, that is, its expected share of all gradient updates. Under our batch rule this weight decreases with the session's position, because every later session reuses the earlier ones: in one example the first session had a share of 0.19 and the last 0.09. Weighted retrain was equal to or worse than plain retrain in all 8 settings we tried (takeover and $\alpha{=}0.3$; $k{=}1$ and $4$; orders A\,B\,C and C\,B\,A), and it strengthened the bias toward the user who corrects first in each round. Replay sees early sessions more often in total but late sessions last, and it is the second property that matters.

\section{Fading of a habit: attention and partial users}
\label{app:fading}

We vary the share of remaining users who prefer the gap H did not (0, 0.25, 0.5, 0.75 or 1), the number of remaining users (2, 4 or 6) and the probability that a user notices a violation (1.0, 0.7 or 0.4); 10 seeds.

\textbf{Attention.} When users notice only 40\% of violations, slightly more of H's gap habit remains after 4 rounds ($+0.07$ compared with full attention; $p{=}0.08$ to $0.11$ with 10 seeds). A noticing probability of 0.7 makes no difference.

\textbf{Partial users.} With partial users at $\alpha{=}0.3$, H's gap habit starts smaller ($+0.17$ instead of $+0.66$) and fades fully within 2 rounds at full attention ($+0.005$ after 8 rounds). Lower attention leaves a small residue ($+0.056$ at a noticing probability of 0.4 after 4 rounds, $p{=}0.002$). Deleting H's corrections from the buffer helps when three quarters of the remaining users oppose H's gap ($-0.049$, $p{=}0.037$).

\textbf{Disagreement among the remaining users.} When at most half of the remaining users oppose H's gap, the habit fades to zero with takeover users whether or not H's corrections are deleted from the buffer.

\section{Attribution methods and detailed results}
\label{app:attribution}

\textbf{Methods.} We compare four ways to estimate a user's influence $I_v$ from the log.
\begin{enumerate}
  \item \emph{Remove the user by replay}: replay without $v$ and measure the change in the behaviour.
  \item \emph{Remove one session at a time}: replay without each session in turn and add up the changes for each user.
  \item \emph{A quick estimate without retraining}: for each logged update, multiply the stored change in the network's weights by the direction in which the weights would have to move to change the behaviour, and add these up for each user. This is a first-order estimate in the style of TracIn.
  \item \emph{Counting}: count each user's logged corrections that push the behaviour, with a plus or minus sign for the direction (the logged action minus the robot's action at that state, along the behaviour's axis). This uses the log's records only and needs no training.
\end{enumerate}
We also report a random ranking.

\textbf{Results.} Mean Spearman rank correlation with the true influence, over 60 groups of users and seeds:

\begin{center}\footnotesize
\begin{tabular}{lccccc}
  \toprule
  users & 1 remove user & 2 remove session & 3 quick estimate & 4 counting & random \\
  \midrule
  takeover & 0.25 [0.19, 0.31] & 0.19 & 0.11 & 0.24 & $-0.01$ \\
  partial, $\alpha{=}0.3$ & 0.32 [0.25, 0.38] & 0.31 & 0.18 & 0.25 & 0.05 \\
  \bottomrule
\end{tabular}
\end{center}

Removing one session at a time is no better than removing the user and costs 1.7 times as much. The quick estimate is the weakest, and in the harmful-user case its sign flips between takeover and partial users ($-0.30$ and $+0.72$).

\textbf{Per behaviour.} With method 1, the correlation is moderate for the share of low-gap trips (0.56) and close to zero for average drop position (0.12 with takeover users, 0.30 with partial users) and average speed (0.08 to 0.11). Unlike removal, attribution does not get worse with partial users ($+0.07$, $p{=}0.16$).

\textbf{Noise in the true influence.} The true influence of each user comes from one rerun of the deployment without that user, and training noise is large in this task (Section~\ref{sec:limits}). Part of the low correlation may therefore come from noise in this reference; we did not measure how much.

\section{Seeds where the base policy is weak}
\label{app:seeds}

On Delivery2D, $\piz$ succeeds in fewer than 80\% of episodes on 3 of 20 seeds (52\%, 55\% and 56\%); on Arm3, on 4 of 20 (60\% to 78\%). On Arm3 we repeated every analysis with only the 16 seeds where $\piz$ reaches 80\%; no result changed by more than a few hundredths, and no conclusion changed. On Delivery2D we report all 20 seeds.

\end{document}